\documentclass[runningheads]{llncs}
\usepackage[authoryear]{natbib}

\usepackage{hyperref}

\usepackage{paralist} 
\usepackage{booktabs} 
\usepackage{multirow} %
\usepackage{multicol} %
\usepackage{longtable}
\usepackage{todonotes}

\usepackage{amsmath}
\usepackage{booktabs} 
\usepackage{caption} 
\usepackage{subcaption} 
\usepackage{graphicx}
\usepackage{pgfplots}
\usepackage[utf8]{inputenc}
\usepackage{multicol}
\usepackage{hyperref}
\usepackage[inline]{enumitem} 
\definecolor{blue}{HTML}{1F77B4}
\definecolor{orange}{HTML}{FF7F0E}
\definecolor{green}{HTML}{2CA02C}

\pgfplotsset{compat=1.14}

\begin{document}
\title{MultiAzterTest: a Multilingual Analyzer on Multiple Levels of Language for Readability Assessment}
\titlerunning{MultiAzterTest}
%
\author{Kepa Bengoetxea \and 
Itziar Gonzalez-Dios} 


%
\authorrunning{Bengoetxea and Gonzalez-Dios}
%

\institute{	Ixa group, HiTZ center,University of the Basque Country (UPV/EHU)\\
\email{\{kepa.bengoetxea,itziar.gonzalezd\}@ehu.eus}
}
\maketitle              
\begin{abstract}
Readability assessment is the task of determining how difficult or  easy a text is or which level/grade it has. Traditionally, language dependent readability formula have been used, but these formulae take few text characteristics into account. However, Natural Language Processing (NLP) tools that assess the complexity of texts are able to measure more different features and can be adapted to different languages.  In this paper, we present the MultiAzterTest tool: (i) an open source NLP tool which analyzes texts on over 125 measures of cohesion, language, and readability for English, Spanish and Basque, but whose architecture is designed to easily adapt other languages; (ii) readability assessment classifiers that improve the  performance of Coh-Metrix in English, Coh-Metrix-Esp in Spanish and ErreXail in Basque; iii) a web tool.  MultiAzterTest obtains 90.09~\% in accuracy when classifying into three reading levels (elementary, intermediate, and advanced) in English  and  95.50~\% in Basque and 90~\% in Spanish when classifying into two reading levels (simple and complex) using a SMO classifier. Using cross-lingual features, MultiAzterTest also obtains competitive results above all in a complex vs simple distinction. 

\end{abstract}
\section{Introduction}

Readability assessment is a Natural Language Processing (NLP) research line that aims to classify texts according to their degree of complexity. Traditionally, this assessment has been done with readability formulae, but these metrics take into account few linguistic features and are language-dependent. NLP tools, however, allow to capture more linguistic and stylistic features and speed up the process of the calculation. Recently, neural approaches have been proposed but they need big corpora which are unfortunately not publicly available for this task. Moreover, up to now text length is a problem for them  and the computing requirements are  expensive. Readability assessment can also be considered as a text classification problem \citep{mironczuk2018recent}.

In this paper, we present  MultiAzterTest, the multilingual version of AzterTest \citep{aztertest}, an open-source NLP based tool and web service for text stylometrics and readability assessment. MultiAzterTest is a computational tool that produces indices of the linguistic and discourse representations of a text. MultiAzterTest is based on a standard formalism and  analyzes texts in Basque, Spanish and English, which are typologically diverse languages. Exactly, MultiAzterTest  analyzes 125 linguistic and stylistic features in Basque, 141 in Spanish and 163 in English. Examples of these features are word frequencies, sentence lengths, vocabulary levels, argument overlaps or use of connective devices. Based on the features, SVM classifier is used to classify the text. We use the SMO classifier because it is  one of the most-used and successful  classifier for readability prediction \citep{benjamin2012reconstructing}. Moreover, we compare our results with the state of the art tool's such as Coh-Metrix \citep{graesser2011cohMetrix} in English, Coh-Metrix-Esp \citep{cohmetrixES} in Spanish and ErreXail \citep{errexail} in Basque which are well-known computational tools which analyze  linguistic and discourse indices.

Following we detail the research questions we will address in this paper:
\begin{itemize}
\item (RQ1) Can be features easily adapted to different languages?
\item (RQ2) Which is the impact of preprocessing tools?
\item (RQ3) Which features are the most predictive? Are they shared in different languages?
\item (RQ4) Do the features or feature groups have the same accuracy in different languages?
\item (RQ5) Is it possible to find common features for all the languages and apply them competitively? That is, is cross-lingual readability competitive?
\end{itemize}

The use of MultiAzterTest is not limited to readability assessment, it can be used for  text analysis, profiling or stylometrics. Text analysis has been used in other research areas such as textbook analysis \citep{aguirregoitia2020clil}, fake news detection and classification \citep{choudhary2020linguistic}, authorship attribution \citep{hou2020robust},  misogyny  identification \citep{fersini2020profiling}, register analysis \citep{argamon2019computational}, analysis of literature \citep{melka2019stylometric},  plagiarism detection \citep{foltynek2019academic}, analysis of the  writing differences of  women and men  \citep{cocciu2018gender}, analysis of the narratives in schizophrenia \citep{willits2018evidence} or detection of dementia  \citep{cunha2015coh}. 

The main contribution to the field of feature analysis of this work is the effort of  integrating multilingual tools to obtain an open-source NLP based tool  for text stylometrics and readability assessment. We also contribute by improving the state-of-the-art results in readability assessment in monolingual setting in Basque, Spanish and English and show that multilingual approaches are  competitive. Moreover, we present a web service.

This paper is structured as follows: in Section \ref{sec:relatedWork} we introduce the related work,  in Section \ref{sec:MultiAzterTest} we present  MultiAzterTest, which we evaluate in Section \ref{sec:evaluation}. In Section \ref{sec:discussion} we discuss our approach and finally, we conclude and outline the future work in Section \ref{sec:conclusions}.

\section{Related work}
\label{sec:relatedWork}

Linguistic profiling is getting more fine-grained and offers more linguistic and stylistic features due to the improvements in NLP processing tools and recent resources. For example, \cite{brunato2020profiling} present  Profiling–UD, a multilingual text analysis tool based on the Universal Dependencies framework. Profiling–UD analyzes more than 130 features from different linguistic levels and more than 50 languages. There are also packages like {\it stylo} that facilitate the stylometric analysis in R \citep{eder2016stylometry} and tutorials for the analysis \citep{Cuentapalabras}.

Traditionally,  text complexity analysis  has been done readability  formulae such as  Flesch \citep{flesch1948new},  Dale-Chall \citep{DaleChallRevisited95},  the indexes Gunning FOG \citep{gunning1968technique} or Simple Measure Of Gobbledygook (SMOG) grade \citep{mc1969smog}. In general, these formulae are based  on  raw features such as word and sentence length, vocabulary lists and frequencies and give a score to classify texts and are language-dependent (most of them only for English). These measures have been widely used to assess reading materials in education. However, NLP based tools have proved that these formulae are not reliable when assessing the levels of the texts \citep{si2001statistical,petersen2009machine,feng2010comparisonRedability}. Moreover, the information offered by these traditional formulae is insufficient,  since they do not detect slight changes in aspects such as coherence and cohesion of the texts \citep{graesser2011cohMetrix}. 

Computational tools, however, can focus on the quantitative dimension of texts and  word length, frequencies, incidences of grammar structures, semantic information, cohesive devices are used to assess linguistic complexity or for  authorship  verification  and  recognition. For English texts, Coh-Metrix 3.0 \citep{graesser2011cohMetrix} analyzes  110 measures at different linguistic levels (descriptive, readability, text easability principal components scores, word   information,  lexical diversity, syntactic complexity, syntactic pattern density, latent  semantic analysis, connectives, referential  cohesion and situation model) in its free version.  Other features that have  been taken into account  are  entities, lexical chains and coreference   \citep{feng2010comparisonRedability}.  \cite{vajjala2018onestopenglish}    use generic text classification features (n-grams, syntactic production rules and dependency relations) and features used in readability assessment. Finally, AzterTest \citep{aztertest}  measures 153 features, which, moreover, include word frequencies and vocabulary knowledge. There are also tools such as TAALES that focus on lexical features \citep{kyle2018tool} or TAACO on cohesion \citep{crossley2019tool}. 

Regarding the readability assessment experiments in English,   \cite{vajjala2018onestopenglish}  use a Sequential Minimal Optimization (SMO) classifier with linear kernel and the best result they obtain is 78.13~\% of accuracy using the total features.  \cite{aztertest} use  SMO and SL classifiers trained with the output of the Coh-Metrix and AzterTest and the best results is 82.01~\% of accuracy in 10-fold cross-validation,  90.09~\% in test set when  using the output of AzterTest with the SMO classifier with the most predictive 50 features. All these experiments have been carried out using using the OneStopEnglish corpus, which has three readability levels: elementary, intermediate and advanced.

Coh-Metrix has also been adapted to Spanish \citep{cohmetrixES}.  Coh-Metrix-Esp calculates 45 readability indices (descriptive, readability, lexical diversity, word information, syntactic complexity, syntactic pattern density, connectives and referential  cohesion). The best result in the simple vs complex distinction is also obtained with SMO in the 10-fold cross-validation using all the indices, which is namely a $F$ measure of  0.9.

 ErreXail \citep{errexail} is  a readability assessment system for Basque  and it calculates 94 indices based on global, lexical, morphological, morpho-syntactic, syntactic and pragmatic features. ErreXail obtains its best results when classifying complex vs simple texts in the 10-fold cross-validation with an optimized SMO classifier: 89.90~\% in accuracy with all the features, 90.75~\% with only lexical features and 93.50~\% with a combination of lexical, morphological, morpho-syntactic and syntactic features.

In multilingual approaches that also analyze Basque, English and Spanish, shallow, morphological, syntactic and semantic features have in taken into account \citep{madrazo2020cross}. In this case, different parsers have been used: {\it Freeling} for English, Spanish, French, Catalan, and Italian \citep{padro2010freeling} and {\it Katea} for Basque \citep{bengoetxea2010application}. They obtain the best results using Random Forest as learning model. Considering all the features, the accuracy when classing simple and complex texts  is  95~\% for English,  87~\% for Spanish and 86~\% for Basque.   However, the best result for English is obtained with shallow features, which is an accuracy of 96~\%.

Following, we present the works for other languages that apply linguistic profiling before readability assessment. Coh-Metrix has been adapted to Brazilian Portuguese \citep{scarton2010coh,scarton2010analise}. The Brazilian Portuguese version, Coh-Metrix-Port, adapts 41 metrics and its based on the second version of Coh-Metrix. The metrics included are readability metrics, words and textual information, syntactic information and logical  operators. When using these features for readability assessment, three levels of complexity are distinguished.

Focusing on the simple vs complex classification,  READ–IT  \citep{DellOrletta:2011} analyzes raw, lexical, morpho–syntactic and syntactic features  at sentence and at document level in Italian.  In addition to the previous features,  in  German  language model features are taken into account \citep{hanckeReadabilityGerman12}. In the case of French,   the CEFR levels are assessed \citep{franccois2012ai}. To that end, 406  lexical, syntactic, semantic, and   French as foreign language  features are analyzed. In a bilingual setting for English and Dutch, \cite{de2016all}   present a system that analyzes traditional features,  lexical features, syntactic features, and semantic features. In addition to the classification, they have also experimented with regression in order to  an absolute score for a given text. 

Recently, neural networks have also been used in multilingual readability assessment. To make simple or complex classification,   \cite{madrazo19} present a multiattentive recurrent deep learning architecture for various languages  \cite{madrazo19} and  \cite{schicchi2020deep}  use recurrent neural units and the attention mechanism for Italian.  

Finally, related to text complexity and readability assessment, the task of Complex Word Identification (CWI) is also gaining attention in the NLP community. A proof of that are the shared tasks organized  in SemEval 2016 \citep{paetzold2016semeval}, the  Innovative Use of {NLP} for Building Educational Applications workshop at NAACL-HTL 2018 \citep{yimam-etal-2018-report}, or ALexS 2020 (only for Spanish) at IberLEF  \citep{ortiz2020overview}. These works, however, focus only on words or multiword expressions and Basque has not been analyzed.

\section{Building MultiAzterTest}
\label{sec:MultiAzterTest}

In this section, we describe MultiAzterTest  which is freely available from a public GitHub repository (we will make the urls available when the paper is accepted) and it is licensed under GNU General Public License v3.0. MultiAzterTest is sensitive to cohesion relations, world knowledge, and language and discourse characteristics based on  Graesser et al.  theoretical information  among others \cite{graesser2004coh}. 

MultiAzterTest is the multilingual version of AzterTest \citep{aztertest}.  In this work, we do not only propose the features that best fit in a multilingual readability assessment context, but we also provide an universal architecture based on dependencies that allows  to cover in a broader sense more applications for different text analysis. MultiAzterTest includes the following improvements:

\begin{itemize}
\item  MultiAzterTest analyzes more than 125 linguistic and stylistic features in 3 languages (see table \ref{tab:numberoffeatures})

\item  MultiAzterTest includes the possibility of selecting  more than one parser [NLP-Cube (0.1.0.7) or StanfordNLP (0.2.0)] or adding other parsers for Basque, English and Spanish based on Universal Dependencies (UD). To add a new model, the load and download methods must be added in the NLPCharger class.

\item MultiAzterTest includes the possibility of adding others languages. To that end, the information needed is: 
\begin{itemize}
\item A file containing the language connectors ordered by the following categories (causal, logical, adversative, temporal and conditional connectives)
\item A file containing a list of the irregular verbs
\item A file containing a list of the stopwords
\item A file containing a list of words at each CEFR level
\item A binary file with the FastText  embeddings \citep{mikolov2018advances}
\item A tool to split syllables
Once these resources have been added, the path to load them should be added in the load/download methods of the Connectives, IrregularVerbs, Stopwords, Oxford, Similarity and Word classes respectively.
\end{itemize}

\item MultiAzterTest offers the possibility to select only certain groups of indicators to be calculated by activating checkboxes. In case of not selecting any group of indicators, the default action will be to calculate all indicators. With this option the user can focus on different aspects of language depending on our purpose.

\item MultiAzterTest is separated into multiple classes that separate and order all the methods, making it much easier and more intuitive to find any functionality. The main difference with respect to the original AzterTest lies in the separation of the methods that analyze the text. In MultiAzterTest, classes are divided into lexical units: document, paragraph, sentence and word; where each class contains its corresponding methods.

\item MultiAzterTest improves AzterTest performance, both in runtime and memory usage. AzterTest restricted the maximum number of files to be analyzed simultaneously to 5 if the semantic similarity indicators were selected. At this moment, MultiAzterTest has no limit on the number of files to be analyzed simultaneously.

\end{itemize}

\begin{figure}[!ht]
\centering
\includegraphics[width=0.8\textwidth]{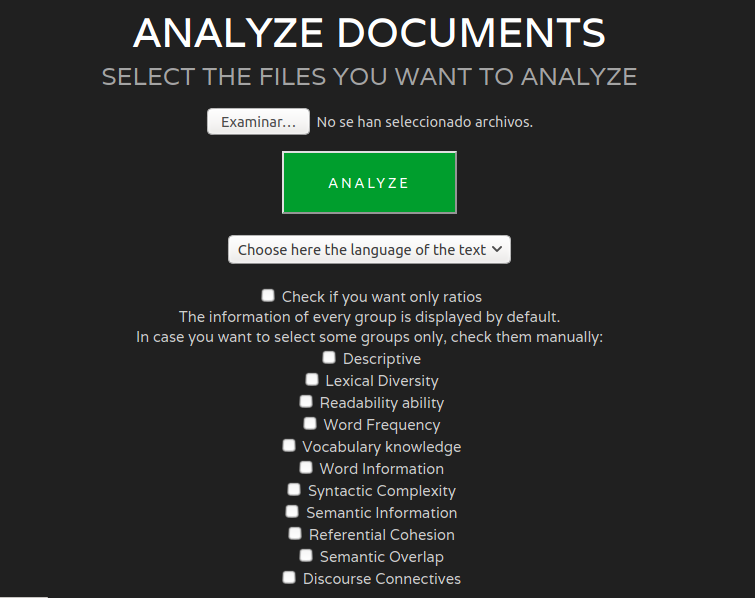}  
\caption{Main Page of the MultiAzterTest Web Tool}
\label{fig:webservice}
\end{figure}

 MultiAzterTest is also available as web service.\footnote{The application can be tested at \url{http://ixa2.si.ehu.eus/aztertest/}} In this service the user can upload a text in Basque, Spanish or English, and will be able to download the output of features together with the complexity level. In Figure \ref{fig:webservice} we show the home page of MultiAzterTest.

Following, we present the tools and resources that we have used. In Subsection \ref{sec:tools} we describe the resources needed for the design and implementation of MultiAzterTest, in Subsection \ref{sec:architecture} we show the architecture of MultiAzterTest and in \ref{sec:features} we introduce the linguistic and stylistic features MultiAzterTest analyzes.

\subsection{Tools and Resources}
\label{sec:tools}

In order to implement the metrics, the raw texts need to be processed and linguistic information needs to be added in the preprocess. Following,  we detail the tools and the resources used by MultiAzterTest.

\begin{itemize}
\item {\bf Multilingual parsing}:  For the automatic analysis of the text, with the aim of easy to adapt MultiAzterTest  to as many languages  as possible, we have decided to use parsers that adopt the Universal Dependencies (UD) formalism. Exactly, we have tested NLP-Cube (0.1.0.7)  \citep{boros2018nlp} and StanfordNLP (0.2.0) \citep{qi2019universal}, that were one of the best systems for English, Basque and Spanish in the CoNLL 2018 Shared Task: Multilingual Parsing from Raw Text to Universal Dependencies \citep{K18-2:2018}. Both parsers carry out the following analysis: segmentation (tokenization and sentence-splitting), lemmatization, PoS tagging and dependency parsing for over 50 languages. 

\item  {\bf Syllable splitting}:  To count the number of syllables each word in a text has, we need to use language-dependent tools.  For English, we have used a syllable splitter based on CMUdict (Carnegie Mellon University Pronouncing Dictionary) \citep{weide2005carnegie}; for Basque, we have used a  rule-based syllabifier  \citep{agirrezabal2012finite}; and, for Spanish we have used the {\it sibilizador} syllable splitter \citep{esSilaba}.

\item  {\bf Stop words removing}: To identify and remove the words with no content,  we have used the stop words lists from the Stopwords ISO \citep{stopwords}, a collection of stopwords for multiple languages. 

\item {\bf Word frequencies}: For word frequency in Spanish and English, we have used wordfreq \citep{robyn-speer},  which provides frequencies of words for 36 languages. wordfreq detects the word frequency  of a word as the logarithm in base 10 of the number of times a word appears per one billion words. A word rated as 3 appears $10^{3}$ times for every $10^{9}$ words, that is, once per million words. As in AzterTest, we have decided to keep words with a value below 4 as rare words. In the case of Basque, we have followed the corpus-based strategy proposed by  \cite{laguntest} since  Wordfreq does not provide frequencies for this language.  

\item {\bf Word levels}: In the case of English, we have added the list provided by the Oxford Learners dictionary, which is  a list of the 3,000 core words at A1-B2 level and an additional 2,000 word  at B2-C1 level \citep{OxfordList}.

\item {\bf Semantic information}: For semantic  information,  we have used wordnets, lexico-semantic resources based on Princeton WordNet  \citep{miller1995wordnet}, which group nouns, verbs, adjectives and adverbs into sets of cognitive synonyms (synsets) to express different concepts and are interlinked by means of conceptual-semantic and lexical relations.  For the texts in English we have used the version included in the NLTK toolkit \citep{steven2006nltk} and for the texts in Basque and Spanish  we have used the 3.0 version of the Multilingual Central Repository (MCR) \citep{gonzalez2012multilingual}. We have used synsets to obtain  the polysemy of lexicon words and the hypernym of verbs and nouns.

\item {\bf Word embeddings}: To calculate semantic similarity,  we have used FastText  embeddings \citep{mikolov2018advances} for English, Spanish and Basque. 

\item  {\bf  List of irregular verbs:} To detect irregular verbs, we have used available lists for English \citep{IrregularVerbsEN} and Spanish \citep{IrregularVerbsES}. 
\item {\bf  List of connectives: } We have  elaborated  a  list  of  connectives  and  its categories for each language, merging linguistic and educational resources from different sources.

\end{itemize}

By using multilingual tools as much as possible, we facilitate the adaptation from one language to other by design (RQ1). In Section \ref{sec:discussion} we discuss on how to adapt MultiAzterTest for other langaguges. Regarding the resources we have gathered, we make all our models publicly  available (when the article is accepted).

\subsection{Architecture}
\label{sec:architecture}

\begin{figure}[!ht]
\centering
\includegraphics[width=0.5\textwidth]{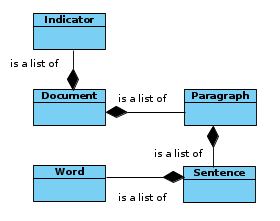}  
\caption{The UML diagram of the data classes}
\label{fig:domainmodel}
\end{figure}

The internal architecture of the MultiAzterTest is based on two kinds of classes: data classes and processing classes. 
In Figure \ref{fig:domainmodel} we present an UML diagram with the data classes.
 
 \subsubsection{Data Classes}
 Data Classes are the basic classes which are used to contain  linguistic data (such as word, sentence, paragraph and document). Any parser that adopts the Universal Dependencies (UD) formalism can be used (such as NLP-Cube or StanfordNLP). Both NLP-Cube and StanfordNLP  carry out the following analysis: segmentation (tokenization and sentence-splitting), lemmatization, PoS tagging and dependency parsing. These parsers load data classes in order to be able to provide to each processing module the right data, and to correctly interpret the module results.  The linguistic classes supported by the current version are:
 \begin{itemize}
\item {\it Indicator} represents all the linguistic and stylistic features calculated during the analysis using the python subclass called defaultdict.  
\item {\it Word} contains the attributes extracted from the parser such as 
index or word number, form, lemma, category, subcategory, morphosyntactic features (case, number, type of subordinated sentence...) and the dependency relation (headword + dependency). In addition, methods related to the characteristics of words are implemented, such as whether it is a personal pronoun, verb, lexical word...
\item {\it Sentence} contains a list of words of the {\it Word} class. This class implements all the methods which calculate the indicators about the syntactic structure of the sentence.
\item {\it Paragraph}  contains a list of sentences known to be as  an independent paragraph.
\item {\it Document}  contains a list of paragraphs that form a complete document to analyze. This class implements the methods which calculate all numbers, means, ratios... by  loading a dictionary of the indicators using the parser information and additional resources or additional lists.

\end{itemize}

 \subsubsection{Processing classes}
The processing classes transform the linguistic classes. The processing classes supported by the current version are:
 \begin{itemize}
\item {\it Printer} contains i) a dictionary of all the indicators and ii) a dictionary of the description of each indicator and different lists such as ignore list of indicators depending of the language, ignore list of counters indicators, ignore list of similarity indicators and lists of groups that classifies the indicators in different linguistic levels such as descriptive, lexical, semantic, syntax, discourse...  This class implements different methods to show the indicators on the screen and generate two different csv files: the first contains the necessary indicators depending on each language and the mode selected by the user. The second file contains those indicators that the classifier uses to predict.
\item {\it NLPCharger} implements the methods which are in charge of  downloading the parser from the web and loading the correct model based on the analyzer and language chosen by the user. In addition, it cleans the text to remove the  strange characters. Finally, NLPCharger transforms the cleaned data into an unified structure (Document, Paragraph, Sentence and Words) using ModelAdapter.
\item {\it  ModelAdapter} contains the chosen parser model  and implements the  method  which transform the parser's output of the document text into an unified structure consisting of document, paragraph, sentence and word. 

\item {\it Predictor} evaluates the complexity level of a text  (elementary, intermediate or advanced in English and complex or simple in Spanish and Basque) using the calculated indicators and the trained model.
\item {\it Resources} represents the different class resources used during the analysis to obtain some of indicators. For example, Oxford class which contains the lists of Oxford vocabulary by level, Connectives class which contains lists of discourse connectives, IrregularVerbs class which contains a list of Irregular Verbs, Stopwords class which contains a list of stopwords, Similarity class which contains the wordembeddings and so on.
\end{itemize}

\subsection{Linguistic and Stylistic Features}
\label{sec:features}

MultiAzterTest calculates linguistic and stylistics features. Linguistic features  are those related to morphology, syntax and semantics while the stylistic features are related to cohesion, vocabulary knowledge. etc.

MultiAzterTest includes different type of scores for each indicator such as absolute numbers, mean, standard deviation, incidence and ratios. Following, we explain these scores:

\begin{itemize}
\item {\bf Absolute numbers:} counts of the occurrences of certain features in the text e.g. the number of verbs in the text.
\item {\bf Incidence:} the incidence as the number of classified units per one thousand words e.g. the incidence score for nouns  computes the number of words that are classified as nouns for a span of 1000 words. 
\item {\bf Ratio:} a ratio score is a relative measure that compares the incidence of one class of units to the incidence of another class of units. For example, noun density ratio is the incidence of nouns divided by the incidence of words. Ratio scores compare two different metrics (classes of units) whereas an incidence score applies to only one metric. 
\item {\bf Mean:} the mean as the sum of the numbers divided by how many numbers are being averaged e.g. the average number of words in each sentence within the text.
\item {\bf Standard deviation:} the standard deviation as a measure of the amount of variation or dispersion of a set of values e.g. the standard deviation of the measure for the mean length of paragraphs within the text.
\end{itemize}

Since the value of some absolute numbers varies greatly depending on the length of the text, we offer the possibility to exclude absolute numbers. Although the raw number of words is usually a predictive feature in readability assessment, it depends on text length and not on its linguistic characteristics. In Table \ref{tab:numberoffeatures} we present the number of features analyzed for each language.

\begin{table}[!htp]
\centering
\begin{tabular}{l c c }
\hline
\textbf{Language}& \textbf{Absolute numbers and ratios} &\textbf{Only ratios} \\
\hline
English & 163 & 116 \\
\hline
Spanish & 141  & 104 \\
\hline
Basque	& 125 & 95 \\
\hline
\end{tabular}
\caption{The number of linguistic and stylistic features in 3 languages}
\label{tab:numberoffeatures}
\end{table}

Following, we present  the list of the features MultiAzterTest analyzes organized by type:

\begin{itemize}
 \item {\bf Descriptive and raw features:}  MultiAzterTest provides descriptive indices to help the user check the output (e.g., to make sure that the numbers make sense) and interpret patterns of data. The extracted indices include  number and ratios of letters,  syllables, lemmas, words, sentences and paragraphs. 
 
 \item {\bf Lexical diversity (only ratios):} These metrics analyze the different and unique words used in the text e.g.  lexical density, densities of nouns, verbs, adjectives and adverbs; simple type-token ratio, content type-token ratio, type-token ratio of nouns, verbs, adjectives and adverbs; lemma simple type-token ratio, lemma content type-token ratio, lemma noun, verb, adjective, and adverb type-token ratio; Honor\'e Lexical Density, Maas Lexical Density, and Measure of Textual Lexical Diversity (MTLD) \citep{mccarthy2005assessment}.

 \item {\bf Classical readability formulae (only ratios):}  The traditional method of assessing texts on difficulty consists of various readability formulas. MultiAzterTest calculate some of the most common formulas such as Flesch readability ease for English \citep{flesch1948new} and Spanish  \citep{fernandez1959medidas} and  Simple Measure Of Gobbledygook (SMOG) grade for English  \citep{mclaughlin1969clearing}. There is no readability formula for Basque.

\item {\bf  Word frequencies:} These metrics are associated to vocabulary acquisition. Using the tools and resources in Subsection \ref{sec:tools},    MultiAzterTest obtains the values of word frequency and  calculates a list of metrics, such as,  minimum word frequency per sentence;  number and incidence of rare words and rare words by each PoS, distinct rare content words; mean of rare content words and distinct rare content words.

\item {\bf Vocabulary knowledge:} The  Common European Framework of Reference for Languages (CEFR) is a framework to describe language ability in six levels. MultiAzterTest calculates number and incidence of words at each CEFR level (only for English).
\item {\bf Word morphological information:} These metrics analyze the form and structure of the words. MultiAzterTest considers number and incidence of each PoS and morphological features of each PoS (e.g. tense, mood, person...), content words and  ratio of proper nouns per nouns. 
\item {\bf Syntax:}  These metrics focus on the govern and the structure of sentences e.g. left embeddedness; mean of descendants and modifiers per noun phrase; propositions, noun phrase and verb phrase per sentence; number and incidence of subordinate clauses, passive and negation.
\item {\bf Semantic information (only ratios):} These metrics take into account the  mean values of polysemy and mean values hypernym values.  
\item {\bf  Semantic overlap (semantic similarity):} This measures provide semantic overlap between sentences or between paragraphs based on FastText word embeddings. MultiAzterTest measures mean and standard deviation of semantic similarity between adjacent sentences, paragraphs and all possible pairs of sentences in a paragraph.
\item  {\bf Referential cohesion (overlaps) (only ratios):} These metrics indicate overlap in content words, exactly mean values of noun, stem, argument and content word overlap.
\item  {\bf Logical cohesion (connectives):} These features show the number and incidence of  causal, logical, adversative, temporal and conditional all and connectives.
 \end{itemize}

\section{Evaluation}
\label{sec:evaluation}

In this section, we present the evaluation of MultiAzterTest in readability assessment. In Subsection \ref{sec:corpus} we describe the corpora we have used in the experiment  and in Subsections \ref{sec:monoresults} and \ref{sec:multiresults} we show the results we have obtained with MultiAzterTest in monolingual and multilingual experiments. We compare MultiAzterTest’s results with the state of the art tools such as Coh-Metrix and AzterTest in English, Coh-Metrix-Esp in Spanish and finally ErreXail in Basque.

\subsection{Corpora}
\label{sec:corpus}

In order to  validate MultiAzterTest, we have used 3 corpora (one for each language). 
For our experimental purposes, we have used a 10-fold cross-validation balanced by level of complexity, where the dataset were partitioned into 10 balanced groups. We have trained 10 times on 9/10 of the labeled data and we have evaluated the performance on the other 1/10 of the data.   We have included the same number of text for each class because the balancing of classes is important, which significantly affects the classifiers’ performance. In the case of English, we have also performed complementary experiments using a train and a test set, since the corpus is bigger. Following we present the details of each corpus.

\begin{itemize}

\item {\bf Basque corpus}: The  Leveled Basque Science Popularisation Corpus (LBSPC)  is composed of 400 texts at 2 levels: simple texts for children and complex texts for adults. The complex texts, henceforth T-comp, are composed by 200 texts (100 articles and 100 analysis) from the {\it Elhuyar aldizkaria}, a  journal about science and technology in Basque. T-comp is meant to be the complex corpus. The simple texts, henceforth T-simp, are composed by 200 texts from {\it ZerNola}, a website to popularize science among children up to 12 years and the texts are articles. In both levels, the texts are from the popular science domain. This corpus was used as  dataset  for readability assessment  in Basque \citep{errexail}. In Table \ref{tab:corporaEU} we present the text, sentence and word number of the dataset together with the mean of sentence length, lexical density and depth of the sentences according to   MultiAzterTest for each complexity level.

\begin{table}[!h]
\centering
\begin{tabular}{c|c|c}
\hline
{\bf  EU} &  {\bf Simple} &  {\bf Complex}  \\ \hline 
  {\bf Texts} &  200 &   200 \\
  {\bf Sentences} & 3118 &  9602 \\
  {\bf Words} &  42586 & 171147 \\
  {\bf Sent len. (m)}& 11.35 &  14.87 \\
  {\bf Lex. den. (m) } &  0.6617 & 0.6567 \\
  {\bf Depth per sent.(m)}   & 4.35  & 5.16   \\  \hline
\end{tabular}
\caption{Information about the Basque corpus obtained by MultiAzterTest (Stanford)}
\label{tab:corporaEU}
\end{table}

\item {\bf Spanish corpus}: The corpus for Spanish is composed of 100 texts at 2 levels: 50 simple texts for children and 50 complex texts for adults. The simple texts are mainly children’s fables while the complex ones are stories for adults. These two corpora were used as  dataset  for readability assessment  in Spanish \citep{cohmetrixES}. In Table \ref{tab:escorpus1} we show the statistics of the corpus.

\begin{table}[!h]
\centering
\begin{tabular}{c|c|c}
\hline
{\bf  ES}& {\bf Simple} & {\bf Complex}  \\ \hline 
   {\bf Texts} & 50 & 50 \\
  {\bf Sentences} &  851 & 2472 \\
  {\bf Words}&  16660 (18602)&  33769 (39704) \\
  {\bf Sent len. (m)}& 20.94& 15.04 \\
  {\bf Lex. den. (m) } &  0.4800& 0.4871 \\
  {\bf Depth per sent.(m)}  & 5.52  & 4.71 \\ \hline

\end{tabular}
\caption{Information about the Spanish corpus obtained by MultiAzterTest (Stanford)}
\label{tab:escorpus1}
\end{table}

\item {\bf English corpus}:  The OneStopEnglish corpus \citep{vajjala2018onestopenglish} compiles newspaper articles aligned at text and sentence level in three  levels (elementary, intermediate, advanced). This corpus has been used for readability assessment \citep{vajjala2018onestopenglish,aztertest}  and it is  available with license CC BY-SA 4.0. The corpus consists of 189 texts, each of them in three versions (567 in total). In Table \ref{tab:corporaEN} we show the detailed information on the corpus.

\begin{table}[!h]
\centering
\begin{tabular}{c|c|c|c}
\hline
   {\bf EN} &  {\bf Elementary} & {\bf Intermediate} &  {\bf Advanced}  \\ \hline 
   {\bf Texts} &  189  &  189  & 189\\
  {\bf Sentences} & 6386 &  7038 & 7675 \\
  {\bf Words}& 102641 (115329)& 130476 (146372)& 158421(177508) \\
  {\bf Sent len.(m)}& 16.49&  19.10& 21.34 \\
  {\bf Lex. den. (m) } &  0.4985& 0.5015& 0.5033 \\
  {\bf Depth per sent.(m)}  & 5.08 &  5.49 & 5.85 \\ \hline
\end{tabular}
\caption{Information about the English corpus obtained by MultiAztertest (Stanford)}
\label{tab:corporaEN}
\end{table}

As a complementary experiment, we have also divided the corpus (in total 567 texts) into 2 non-overlapping datasets (80~\% train and 20~\% test): 456 texts (152 texts for each class) as the training set and 111 texts (37 texts for each class) as the test set. 

\end{itemize}

\subsection{Monolingual Experiments and Results}
\label{sec:monoresults}

In this Section we present the readability assessment experiments in 3 languages (English, Spanish and Basque) and we compare MultiAzterTest's results with the state of the art tools such as Coh-Metrix, Coh-Metrix-Esp  and ErreXail.

In order to classify the texts according to their complexity level,  we have trained Sequential Minimal Optimization (SMO) classifier  \citep{platt1998sequential}  that is included in  WEKA \citep{hall2009weka}. We have tested all the classifiers with the defaults hyperparameters,  without any optimization. SMO  is an optimization technique for solving quadratic optimization problems, which arise during the training of Support Vector Machines (SVM). One of the reason to choose SMO is the high classification accuracy  in the same task reported in the literature \citep{aztertest,cohmetrixES,errexail,hanckeReadabilityGerman12,benjamin2012reconstructing}. As we mentioned, to evaluate the classifier we have used the 10-fold cross-validation balanced by level of complexity and we report our results with the accuracy. In the case of English, we also have carried out a complementary experiment with a train and a test set.

We have carried out two types of experiments related to our research questions (RQ2, RQ3, RQ4). In the first experiment (related to  questions RQ2 and RQ3), we have tested the the preprocessing step  using different parsers  and the suitability of features using all features (absolute numbers  and ratios) or only ratios (features  based on incidence, ratio, mean and standard deviation).

To that end, we have tested the following four preprocessing  configurations of MultiAzterTest for the 3 languages with all the features and only ratios:
\begin{itemize}
\item {\bf MultiAzterTest Stanford all (MAzt-Sta-a):} we have taken into account all the features (absolute numbers and ratios) using Stanford preprocessing tool.
\item {\bf MultiAzterTest Stanford ratios (MAzt-Sta-r):} we have taken into account only ratios (features  based on incidence, ratio, mean and standard deviation) using Stanford preprocessing tool.
\item {\bf MultiAzterTest NLPCube all (MAzt-Cub-a):} we have taken into account all the features (absolute numbers and ratios) using NLPCube preprocessing tool.
\item {\bf MultiAzterTest NLPCube ratios (MAzt-Cub-r):} we have taken into account only ratios (features  based on incidence, ratio, mean and standard deviation) using NLPCube preprocessing tool.
\end{itemize}

Moreover, for the feature selection, we have selected the 75, 50 and 25 most relevant features according to Weka's Information gain (InfoGain) with the aim of finding possible combinations of features.

Finally, to compare our results, we have rerun  Coh-Metrix in English, Coh-Metrix-Esp in Spanish and ErreXail in Basque. These are our baselines.

In the second experiment, we have carried out a linguistically motivated selection of features in relation with research question (RQ4).  And, to that end, we have tested the classifier taking into account the groups of features presented in Subsection \ref{sec:features}  (we also give the result of all the features as a reference). In this experiment, we have only tested the results with the Stanford parser and evaluate them for each language in the 10-fold cross-validation. 

In the following subsections, we present the results for each language and each experiment.

\subsubsection{Results for Basque}
\label{sec:resultsEU}

In Table~\ref{tab:basqueresults} we present  the accuracy  of the SMO classifiers when using the output of ErreXail, MultiAzterTest Stanford ratios (MAzt-Sta-r), MultiAzterTest Stanford all (MAzt-Sta-a), MultiAzterTest NLPCube ratios (MAzt-Cub-r) and MultiAzterTest NLPCube all (MAzt-Cub-a) and using different features (Feat. number).

\begin{longtable}{c|l|l}
    \hline
      {\bf Tool/Configuration} & {\bf Feat. number}  & {\bf Accuracy} \\
        \hline 
        \endfirsthead
        \hline
      {\bf Tool/Configuration} & {\bf Feat. number}  & {\bf Accuracy} \\
        \hline 
        \endhead
        \multicolumn{3}{c}{(continued on next page)}
        \endfoot
        \endlastfoot
 
\multirow{4}{*}{ErreXail}&All&86.50 \\
&75&86.00 \\
&50&\textbf{89.50} \\
&25&88.00 \\ 
\hline
\multirow{4}{*}{MAzt-Cub-a}&All&91.00  \\
&75&92.00 \\
&50&93.00 \\
&25&\textbf{94.50} \\
\hline
\multirow{4}{*}{MAzt-Cub-r}&All&92.00  \\
&75&90.50 \\
&50&94.00 \\
&25&\textbf{95.50} \\
\hline
\multirow{4}{*}{MAzt-Sta-a}&All&94.00  \\
&75&94.50  \\
&50&\textbf{95.50}  \\
&25&95.00  \\
\hline
\multirow{4}{*}{MAzt-Sta-r}&All&94.50  \\
&75&94.00 \\
&50&\textbf{95.00}  \\
&25&94.00 \\
\hline
\caption{Classification results of the 2 level readability assessment experiment (cross-validation)} 
\label{tab:basqueresults}
\end{longtable}

When classifying simple vs complex texts  with {\it T-simp} and  {\it T-comp} corpora ErreXail's accuracy  is 86.50~\% taking all the features into account and 89.50~\% using the best 50 features. MultiAzterTest's best results are 95.50~\% in accuracy with the best 50 predictive features in MultiAzterTest-stanford-all  and with the best 25 predictive features in MultiAzterTest-cube-ratios. Therefore, in this scenario, MultiAzterTest outperforms in 6 points the best result of ErreXail.

Analyzing the impact of preprocessing tools when both classifiers used all the features or all the ratios (RQ2),  we can see that the Stanford preprocessor output helps the classifier better predict 2-3 points of the NLPCube preprocessor output. However, this difference is smaller when we test different sets of attributes selected by InfoGain. One of the reasons is to find in the results obtained in the CoNLL 2018 Shared Task \citep{K18-2:2018}  where Stanford's results were better than the results obtained by NLPCube. In the dependency parsing task, Stanford scored on the UD Basque Dependency Treebank 82.75 in LAS while NLPCube scored 81.53 in LAS.

Regarding RQ3, in Table  \ref{tab:basqueTOP10}  we present the 10 most predictive features  in Basque according to InfoGain using all the features (left column) and only ratios (right column). In both columns all the features in the top are ratios and two of them belong to the syntactic group, and the rest of them belong to descriptive group where measures related to lengths seem to be important.

\begin{longtable}{p{6cm}|p{6cm}}
    \hline
       {\bf All} & {\bf Ratios}   \\
        \hline 
        \endfirsthead
        \hline
      {\bf All} & {\bf Ratios}   \\
        \hline 
        \endhead
        \multicolumn{2}{c}{(continued on next page)}
        \endfoot
        \endlastfoot

Lemma length (mean) &	Lemma length (mean) \\
Noun phrases per sentence (mean) &	Noun phrases per sentence (mean) \\
Sentence length without stopwords (mean) &	Sentence length without stopwords (mean) \\
Word length (mean) &Word length (mean)\\
Number of sentences (incidence)	 & Sentence length (mean) \\
Sentence length (mean) &	Number of sentences (incidence)	\\
Word length  without stopword (mean) &	Word length  without stopword (mean) \\
Number of paragraphs (incidence)		& Number of paragraphs (incidence) \\
Number of syllable per word (sd) 	& Number of syllable per word (sd)  \\
Depth of the subordinates per sentence (mean) &	Depth of the subordinates per sentence (mean) \\

\hline
\caption{10 most predictive features in Basque} 
\label{tab:basqueTOP10}
\end{longtable}

In Table~\ref{tab:ratiosbygroupEU} we present the results of the SMO classifier for each specific group of linguistic features, sorted by accuracy (RQ4).  Using all the metrics and ratios of the descriptive group obtain an accuracy of 93~\% and 93.5~\% respectively. This implies that only using the information of the numbers of letters,  syllables, lemmas, words, sentences and paragraphs, the fall of accuracy is one point with regard to the use of all available features in MultiAzterTest. That is, these results are only one point worse than the results with all the features, which means that they are competitive on their own.

\begin{longtable}{c|l|l}
    \hline
       {\bf Feature Group} & {\bf Configuration}  & {\bf Accuracy} \\
        \hline 
        \endfirsthead
        \hline
      {\bf Feature Group} & {\bf Configuration}  & {\bf Accuracy} \\
        \hline 
        \endhead
        \multicolumn{3}{c}{(continued on next page)}
        \endfoot
        \endlastfoot
\multirow{2}{*}{All groups} &All&94.00\\
  &Ratios&94.50\\ \hline 
\multirow{2}{*}{ Description} &All& {\bf 93.00}\\
 &Ratios& {\bf 93.50}\\ \hline

 
\multirow{2}{*}{ Syntactic} &All&82.00\\
 &Ratios&84.00\\ \hline
 
\multirow{2}{*}{Word information} &All&80.50\\
  &Ratios&81.00\\ \hline

{Semantic overlap} &Ratios&78.00\\ \hline

{Lexical diversity}  &Ratios&75.00\\ \hline

\multirow{2}{*}{Discourse connectives} &All&74.50\\
 &Ratios&73.50\\  \hline

\multirow{2}{*}{ Word frequencies}  &All&67.50\\ 
  &Ratios&64.00\\ \hline
  
{Referential cohesion} &Ratios&60.50\\ \hline

{Semantic information} &Ratios&57.50\\ \hline

\hline
\caption{The results of the SMO classifier with specific linguistic  features (Multiaztertest's all/only ratios) in Basque} 
\label{tab:ratiosbygroupEU}
\end{longtable}

By analyzing the rest of the groups by ranges, the syntactic and word information groups are above 80~\%. The use of syntactic (left embeddedness, descendants and modifiers per noun phrase, propositions, noun phrase and verb phrase, subordinate clauses, passive and negation) and morphological (e.g. features of each PoS, content words and proper nouns,...) information seem to help the classifier very well using the Basque corpus.
There are three groups between 70~\% and 79~\% which are semantic overlap (semantic similarity between sentences and paragraphs), lexical diversity (how many different lexical words there are in a text) and discourse connectives (the number of causal, logical, adversative, temporal and conditional connectives). 
Between 60~\% and 69~\% there are two groups, such as, word frequency (e.g. rare words by each PoS, distinct rare content words,...) and referential cohesion  (noun, stem, argument and content word overlap between sentences and paragraphs).  And, finally, the worst result is obtained  by the semantic information with an accuracy of 57.5~\%. This group involves polysemy and hypernym values in Basque WordNet \citep{pociello2011methodology}.

\subsubsection{Results for Spanish}
\label{sec:resultsES}

In Table~\ref{tab:sparesults} we present the accuracy of the classifiers (Tool/Configuration) for each tool and using different features (Feat. number).

\begin{longtable}{c|l|l}
    \hline
       {\bf Tool/Configuration} & {\bf Feat. number}  & {\bf Accuracy} \\
        \hline 
        \endfirsthead
        \hline
      {\bf Tool/Configuration} & {\bf Feat. number}  & {\bf Accuracy} \\
        \hline 
        \endhead
        \multicolumn{3}{c}{(continued on next page)}
        \endfoot
        \endlastfoot
\multirow{2}{*}{Coh-Metrix-Esp} & All (45) & \textbf{87.00} \\
&25&83.00 \\ 
\hline
 \multirow{4}{*}{MAzt-Cub-a}&All&88.00  \\ 
&75&\textbf{89.00}   \\ 
&50&86.00   \\  
&25&84.00   \\
\hline
\multirow{4}{*}{MAzt-Cub-r}&All&85.00   \\ 
&75&\textbf{89.00}   \\ 
&50&86.00  \\ 
&25&88.00 \\
\hline
\multirow{4}{*}{MAzt-Sta-a}&All&\textbf{90.00}  \\
&75&87.00  \\ 
&50&85.00  \\ 
&25&86.00   \\ 
\hline
\multirow{4}{*}{MAzt-Sta-r}&All&\textbf{89.00}   \\ 
&75&86.00   \\ 
&50&87.00  \\ 
&25&85.00   \\ 
\hline
\caption{Classification results of the 2 level readability assessment experiment (cross-validation)} 
\label{tab:sparesults}
\end{longtable}

The first line in Table~\ref{tab:sparesults} shows the accuracy of 87~\% obtained using Coh-Metrix-Esp with all the features (45 features)  using the SMO algorithm in the 10-fold cross-validation. MultiAztertest's  best results are obtained with the Stanford preprocessing tool and using all the features (MultiAzterTest-sta-all) whose accuracy is 90~\% (3 points higher than the best result of Coh-Metrix-Esp).

When making the selection of characteristics using InfoGain, we do not obtain any improvement with respect to Coh-Metrix-Esp and the MultiAzterTest-sta configuration. On the other hand, in the MultiAzterTest-cub configuration,  the best result is an accuracy of 89~\%  using a reduced set of 75 features.

Analyzing the impact of preprocessing tools when both classifiers used all the features or all the ratios (RQ2),  as with the Basque language, better results are obtained when MultiAzterTest uses the Stanford preprocessor output than when using NLPCube. The SMO classifier predicts 2-4 points better than the output of the NLPCube. Indeed, in the CoNLL 2018 Shared Task, on dependency parsing task, Stanford scored on the UD Spanish AnCora Treebank 90.47 in LAS while NLPCube scored 89.06.

Regarding RQ3, in Table \ref{tab:spanishTOP10} we present the 10 most predictive features in Spanish. Contrary to Basque, we see that the raw numbers are quite predictive. Moreover, InfoGain selects a wider variety of features when  using all features or with ratios only. 

\begin{longtable}{p{6cm}|p{6cm}}
    \hline
       {\bf All} & {\bf Ratios}   \\
        \hline 
        \endfirsthead
        \hline
      {\bf All} & {\bf Ratios}   \\
        \hline 
        \endhead
        \multicolumn{2}{c}{(continued on next page)}
        \endfoot
        \endlastfoot
Number of different rare words &	Verb phrases per sentence (mean) \\
Number of sentences &	Distinct rare words (mean)   \\
Number of rare words	& Number of  different rare words (incidence) \\
Number of nouns &	Similarity adjacent paragraphs (mean) \\
Number of lexical words &	Similarity adjacent paragraphs (sd) \\
Number of words with punctuation marks &	Sentences length without stopwords (mean) \\
Number of different forms	& Similarity   paragraphs (mean) \\
Number of logical connectives	& Lemma TTR \\
Number of  words &	Sentence length (mean) \\
Number of total propositions &	Number of sentences (incidence) \\

\hline
\caption{10 most predictive features in Spanish} 
\label{tab:spanishTOP10}
\end{longtable}

Regarding the second experiment (RQ4), in Table \ref{tab:ratiosbygroupES} we  show that the best result is obtained when using all the metrics of the descriptive group, which  obtain an accuracy of 88~\%. This implies that using the descriptive features in their own the accuracy falls only two points when comparing to the use of all available features in MultiAzterTest.

\begin{longtable}{c|l|l}
    \hline
       {\bf Feature Group} & {\bf Configuration}  & {\bf Accuracy} \\
        \hline 
        \endfirsthead
        \hline
      {\bf Feature Group} & {\bf Configuration}  & {\bf Accuracy} \\
        \hline 
        \endhead
        \multicolumn{3}{c}{(continued on next page)}
        \endfoot
        \endlastfoot
 \multirow{2}{*}{ All groups} &All&90.00\\
 &Ratios&89.00\\ \hline
 \multirow{2}{*} {Description} &All& {\bf 88.00}\\
&Ratios&{\bf 85.00}\\ \hline
{Lexical Diversity} &Ratios&83.00\\ \hline
 \multirow{2}{*}{Syntactic} &All&82.00\\
 &Ratios&79.00\\\hline
\multirow{2}{*}{Word frequencies}  &All&81.00\\
  &Ratios&76.00\\ \hline
  \multirow{2}{*}{Discourse connectives} &All&78.00\\
 &Ratios&63.00\\ \hline
 \multirow{2}{*}{ Word information} &All&77.00\\
 &Ratios&66.00\\ \hline
  {Semantic overlap} & Ratios&75.00\\ \hline
  {Semantic information} &Ratios&70.00\\ \hline
 {Referential cohesion} &Ratios&65.00\\ \hline
 {Readability} &Ratios&51.00\\ \hline
\hline
\caption{The results of the SMO classifier with specific linguistic  features (Multiaztertest's all/only ratios) in Spanish} 
\label{tab:ratiosbygroupES}
\end{longtable}

There are other 3 groups that are above 80~\%: the group of lexical diversity, syntactic and word frequency metrics. The groups of discourse connectives, word information, semantic overlap and semantic information obtain an accuracy between 70~\% and 79~\%.  Between 60~\% and 69~\% there is the referential cohesion (noun, stem, argument and content word overlap between sentences and paragraphs).   And, finally, the worst result is obtained using readability group with an accuracy of 51~\% involving only Flesh \citep{flesch1979} readability formulae, which also proves that traditional readability metrics are not reliable.

\subsubsection{Results for English}
\label{sec:resultsEN}

In Table~\ref{tab:en-results} we show the accuracy  of the classifiers (Tool/Configuration) for each tool and using different features (Feat. number). Let us recall that in the case of English we  classify three levels of complexity and that we evaluate in  cross-validation (c) and in a test set (t).

\begin{longtable}{c|l|l|l}
    \hline
       {\bf Tool/Configuration} & {\bf Feat. number}  & {\bf Accuracy (c)} & {\bf Accuracy (t)} \\
        \hline 
        \endfirsthead
        \hline
      {\bf Tool/Configuration} & {\bf Feat. number}  & {\bf Accuracy (c)} & {\bf Accuracy (t)} \\
        \hline 
        \endhead
        \multicolumn{4}{c}{(continued on next page)}
        \endfoot
        \endlastfoot
\multirow{5}{*}{Coh-Metrix} & All & 76.31 & 81.08  \\
& 75  & 77.63 & 82.88  \\
& 50  & 76.75 & 86.48  \\
& 25  & 78.72 & 85.58  \\
\hline 
\hline 
\multirow{6}{*}{MAzt-sta-a}&All&\textbf{82.01} & \textbf{90.09} \\
&75&80.70 &87.38 \\
&50&81.35&90.09 \\
&25&80.48&86.48 \\ 
\hline

\multirow{5}{*}{MAzt-sta-r}&All&\textbf{82.01} &\textbf{86.48} \\
&75&80.04 &88.28 \\
&50&76.09&87.38 \\
&25&76.97& 82.88\\ 
\hline

\multirow{6}{*}{MAzt-cub-a}&All &79.82 &88.28 \\
&75&79.82 &88.28 \\
&50&81.35&89.18 \\
&25&80.48&89.18 \\ 
\hline

\multirow{5}{*}{MAzt-cub-r}&All&80.04 &84.68 \\
&75&81.14 &85.58 \\
&50&75.65&85.58 \\
&25&76.31&83.78 \\
\hline

\caption{Classification results of the 3 level readability assessment experiment (cross-validation)} 
\label{tab:en-results}
\end{longtable}

In the first group of rows we test Coh-Metrix's output obtaining an accuracy value of 76.31~\% in the 10-fold cross-validation when we employ all the features. But we obtain the best accuracy value of 78.72~\% in the cross-validation using a reduced set of 25 features.

In the rest of the row groups we test our four configurations of MultiAzterTest. Regarding MultiAzterTest results, when we employ both Stanford with all the features (MultiAzterTest-Sta-a) and only ratios (MultiAzterTest-Sta-r) we obtain the best accuracy value of 82.01~\% in the 10-fold cross-validation. However, in the test set a better value of 90.09~\% is obtained when using all the features.

When making the selection of features using InfoGain, we do not obtain any improvement with the MultiAzterTest-sta configuration. On the other hand, in the MultiAzterTest-cub configuration, we obtain the best result:  81.35~\% in accuracy in the 10-fold cross-validation using a reduced set of 50 features.

Analyzing the impact of preprocessing tools when both classifiers used all the features or all the ratios (RQ2),  as with the Basque and Spanish languages,  better results are obtained when MultiAzterTest uses the Stanford preprocessor output. The SMO classifier predicts 2 points better than the output of the NLPCube. As with Basque and Spanish Stanford's results were better than the results obtained by NLPCube in the CoNLL 2018 Shared Task: Stanford scored on the UD English Web Treebank 83.87 in LAS while NLPCube scored 82.79 in LAS on dependency parsing task.

In respect to the 10 most predictive features (RQ3), in English, when using all the features raw numbers and ratios are mixed. Among these features, we see the importance of sentence length, and lexical features (the rare words, content words, Honor\'e, Maas).

\begin{longtable}{p{6cm}|p{6cm}}
    \hline
       {\bf All} & {\bf Ratios}   \\
        \hline 
        \endfirsthead
        \hline
      {\bf All} & {\bf Ratios}   \\
        \hline 
        \endhead
        \multicolumn{2}{c}{(continued on next page)}
        \endfoot
        \endlastfoot
      
Number of different rare words &	Sentence length (std) \\
Number of different forms	& Sentence length without stopwords (std) \\
Number of rare verbs &	Distinct rare words (mean) \\
Sentence length (std) &	Number of different rare words (incidence) \\
Number of  words &	Honor\'e \\
Number of content words (not a1 to c1 words) &	Number of rare verbs (incidence) \\
Number of words with punctuation marks &	Sentence length without stopwords (mean) \\
Number of  rare words &	Sentence length (mean) \\
Sentence length without stopwords (std) &	Number of  sentences (incidence) \\
Number of lexical words &	Maas \\
\hline
\caption{10 most predictive features in English} 
\label{tab:englishTOP10}
\end{longtable}

Regarding the second experiment (RQ4), in  Table~\ref{tab:ratiosbygroupEN} we  show the results of the SMO classifier for each specific group of linguistic features, sorted by accuracy.

\begin{longtable}{c|l|l}
    \hline
       {\bf Feature Group} & {\bf Configuration}  & {\bf Accuracy} \\
        \hline 
        \endfirsthead
        \hline
      {\bf Feature Group} & {\bf Configuration}  & {\bf Accuracy} \\
        \hline 
        \endhead
        \multicolumn{3}{c}{(continued on next page)}
        \endfoot
        \endlastfoot
\multirow{2}{*}{ All groups}&All&82.01\\
 &Ratios&82.01\\ \hline
\multirow{2}{*}{Description} &All& {\bf 76.75}\\
 &Ratios& {\bf 69.29}\\ \hline
\multirow{2}{*}{Syntactic complexity} &All&72.80\\
 &All&62.28\\ \hline
\multirow{2}{*}{Word frequency}  &All&72.36\\
&Ratios&65.57\\ \hline
\multirow{2}{*}{Vocabulary knowledge} &All&69.29\\
 &Ratios&55.48\\ \hline
\multirow{2}{*}{Word information} &All&68.42\\
  &Ratios&57.45\\ \hline
{Lexical diversity} &Ratios&64.25\\ \hline
\multirow{2}{*}{Discourse connectives} &All&63.81\\
 &Ratios&39.47\\ 
\hline
{Referential cohesion} &Ratios&54.82\\ \hline
{Semantic information} &Ratios&50.21\\ \hline
{Readability}&Ratios&49.56\\ \hline
{Semantic overlap} &Ratios&37.50\\ \hline
\hline
\caption{The results of the SMO classifier with specific linguistic  features (MultiAzterTest's all/only ratios) in English} 
\label{tab:ratiosbygroupEN}
\end{longtable}

The best result is obtained with the group of all descriptive features, obtaining an accuracy of 76.75~\%. The  descriptive features are the best in the three languages (English, Spanish and Basque).

There are others 2 groups that are above 70~\%, the group of syntactic and word frequency metrics. The syntactic information also helps the classifier very well in all the languages (Spanish and Basque), while word frequency helps only very well in Spanish.

Also, the vocabulary knowledge,  word information, lexical diversity and discourse connectives groups  obtain an accuracy between 60~\% and 69~\%.  Between 50~\% and 59~\% there are the referential cohesion and semantic information.   Finally, the worst results are obtained using readability information group of metrics with an accuracy of 51~\% involving only  the Flesh \citep{flesch1979} readability formulae, which proves again the unreliability of classical readability formulae, and semantic overlap.

\vspace{0.5cm} 
\noindent Comparing  our findings to   \cite{madrazo2020cross}, they show that syntactic features are the ones with the most predictive power, together with morphological features and followed by shallow features. Semantic features were ones that achieve the least average accuracy. We obtain similar results, the descriptive or shallow features group is the best in all the languages (morphology-rich as Basque and not morphology-rich as English). Syntactic features are in second position in Basque and English, and third position in Spanish. Word information or morphological information is in the third position in Basque (a morphology-rich language) and sixth and fifth position in Spanish and English, respectively. In our case too, semantic information and semantic overlap are in the tail of group of features in all the languages.  

\cite{madrazo2020cross} also analyzed the effect that language has on the accuracy of a model. In fact, all models achieve consistently better accuracy for texts in English or Catalan as opposed to Basque. But in our case we obtain better results in Basque than Spanish (in the same 2 levels of complexity) or English (3 levels of complexity).

\subsection{Cross-lingual Experiments and Results}
\label{sec:multiresults}

In Subsection \ref{sec:monoresults} we have seen that MultiAzterTest gets competitive results and outperforms the state-of-the art systems in monolingual settings. In this section we test whether a common set of features in the three languages can be predictive. 

To extract the common cross-lingual features, we have created an intersection of features for the three languages considering all features and ratios only. We present these features in Table \ref{tab:intersection}. 

\begin{longtable}{  p{6cm} |  p{6cm}} 
    \hline
     {\bf All Features}  & {\bf Ratios}\\
        \hline 
        \endfirsthead
        \hline
     {\bf All Features}  & {\bf Ratios}\\
        \hline 
        \endhead
        \multicolumn{2}{c}{(continued on next page)}
        \endfoot
        \endlastfoot

 number of nouns, words with and without punctuation, lexical words,  subordinate clauses, different forms, total propositions, verbs,  adjectives, logical connectives and all connectives
  &

\\ 

mean of sentence length with and without stopwords, verb phrases per sentence, noun phrases per sentence,  subordinate depth per sentence,  propositions per sentence, distinct and rare words and  rare words  
&

mean of sentence length with and without stopwords, word length with and without stopwords,  distinct  rare words, rare words, verb phrases per sentence, noun phrases per sentence,  subordinate depth per sentence, propositions per sentence, paragraph similarity by pairs, and  content overlap (all and adjacent)

\\ 
incidence of number of sentences, number of paragraphs  and rare words

& 
incidence of number of sentences, number of paragraphs,  different rare words and infinitive
\\ \hline

 standard deviation of lemma length and word length
& 
 standard deviation of the number of syllables, lemma length, word length with and  without stopwords, sentence length with and without stopwords,   paragraph similarity by pairs and all content overlap  
\\ \hline

&
index of hypernymy, hypernymy in nouns  and hypernymy in verbs; left embeddedness; honor\'e; vttr, simple ttr, content ttr, adjective and adverb lemma ttr and adjective ttr
\\ \hline
\caption{Intersection of the feature selection for the three languages (all and ratios)} 
\label{tab:intersection}
\end{longtable}

In Table \ref{tab:multilingual}, we present the results of the cross-lingual features in each of the languages. In none of the cases, the cross-lingual results outperform the best monolingual, but they are high competitive in Basque and Spanish where a simple vs complex classification is made. 

\begin{longtable}{c|l|l}
    \hline
       {\bf Feature Group} & {\bf Configuration}  & {\bf Accuracy} \\
        \hline 
        \endfirsthead
        \hline
      {\bf Feature Group} & {\bf Configuration}  & {\bf Accuracy} \\
        \hline 
        \endhead
        \multicolumn{3}{c}{(continued on next page)}
        \endfoot
        \endlastfoot
  \multirow{2}{*}{Basque} &All&92.00\\
 &Ratios&91.50\\ \hline
 \multirow{2}{*}{Spanish} &All&88.00\\
 &Ratios&87.00\\ \hline 
  \multirow{2}{*}{English}&All&77.85\\
 &Ratios&76.53\\ \hline
\hline
\caption{The results of the SMO classifier with the multilingual features} 
\label{tab:multilingual}
\end{longtable}

However, although it cannot be directly compared because it has been tested in other dataset, the results for Spanish and Basque are better that those presented by  \cite{madrazo2020cross}. In a simple vs complex distinction, they obtain 85\% of accuracy for Spanish and 81~\% for Basque with Random Forest. However,  when using support vector machines they drop to 77~\% of accuracy for Spanish and 75~\% for Basque. 

\section{Discussion}
\label{sec:discussion}

In this section we discuss our findings, answer the research questions and point out difficulties and challenges.

Regarding the feature adaptability across languages (RQ1), it is easy to add a new language or a new model. By using multilingual tools as much as possible, we facilitate the adaptation from one language to other by design. Following, we explain which should be changed in order to adapt MultiAzterTest to other languages:

The first tool that needs to be adapted is the parser. To that end, we need to modify  the methods download and load of NLPCharger class (including model download and configuration instructions), by only passing the new language as parameter. This way, we  download and load the model that has been trained in that language  based on universal dependencies. With these changes, we can calculate all the metrics that depend on the morphosyntactic characteristics. 

Stopwords are also available for 60 languages \citep{stopwords}. After downloading stopword list, the list needs to be saved as /data/language/StopWords/stopwords.txt  and the load method in the Stopword class modified. 

By only including these 2 resources (model and stopwords), MultiAzterTest obtains the following metrics: 
\begin{itemize}
\item 20 of 22 metrics of the descriptive group. The mean number of syllables (length) in words and its standard deviation would be left out.
\item All the lexical diversity features (20 features)
\item All the referential cohesion features (10 features)
\item All the word information (34 features) and all the syntactic features (26 features). It is true that, due to the linguistic typology of each analyzed language,  some linguistic features cannot be analyzed as they do not exist in the language. For example, voice is not marked in Basque while it is marked in English and Spanish. This lead us to think that linguistic feature design should not only based on mainstream and majority languages but also in languages with other typologies. Moreover, it is sensible only to analyze the linguistic features present in the language and not the ones there are not. That is, it makes no sense to analyze and give the classifier the voice-based features in Basque.
\end{itemize}

But MultiAzterTest uses more multilingual resources such as wordfreq (available in more than 36 languages), fastText word embeddings (available for 157 languages) and WordNet (available for more than 200 languages). Including these 3 resources, MultiAzterTest obtains the following metrics:   
\begin{itemize}
\item All the word frequency features (15 features) 
\item All the semantic overlap features (6 features)
\item All the Word Semantic information features (4 features)
\end{itemize}

Despite this, there are only 3 resources which are not multilingual: the syllable splitter, vocabulary level lists and  the discourse connectives. Without these resources,  24 out of the  163 possible features would not be calculated. 

By adding an available syllable splitter, MultiAzterTest can calculates  readability metrics (Flesh-based metrics and SMOG) and 2 descriptive features (the mean number of syllables (length) in words and its standard deviation). If we add the lists of discourse connectives, MultiAzterTest calculates 10 discourse metrics such as causal connectives and logical connectives incidence. And, finally, if we add CEFR level vocabulary list, MultiAzterTest calculates 10 more vocabulary  knowledge metrics such as incidence score of A1 vocabulary  (per 1000 words) among others.  

We are aware that adapting resource based features is more difficult since the language needs to have the same resource or a similar one. This happens, for instance, with the CEFR level based features. To our knowledge this resource only exists for English, and that is why these high predictive measures have not been analyzed in Basque or Spanish.  In addition to that, some resources do not have the same coverage. That is the case of wordnets. Although we can use the wordnets for Basque and Spanish to measure lexico-semantic information, they are smaller than the English one.

In relation to the impact processing tools (RQ2), as we have seen in our experiments,  it is important to preprocess the texts with a good parser. In all 3 languages, analyzing the impact of preprocesing tools when both classifiers used all the features or all the ratios,  we can see that the output of Stanford help to the SMO classifier to predict 2 or more than 2 points better than the output of the NLPCube. Looking at the results obtained by each parser in the CoNLL 2018 Shared Task, we can say that the result of the parser improves significantly in the task of classifying texts according to their complexity.  Moreover, we cannot forget that the parser model language may label some words incorrectly labeled, which happens more often if  the treebank used to train the chosen model (Stanford or NlpCube) is small.

With respect to the most predictive features  (RQ3), we see that strategies differ in the three languages and in the three datasets we have used. Basque tends for ratios, the importance of raw numbers is higher in Spanish and a mixed tendency is seen in English. Moreover, we see in general that length and vocabulary measures seem to be highly predictive. This should be further analyzed by using other dataset.

If we analyze the prediction ability by feature groups (RQ4),  descriptive features are highly predictive, above all when a simple vs complex classification is made. This lead us to think that for binary classification task and as we have seen with feature selection few features are enough. The question is, however, if we add more readability levels, will other features play a role?  On the other hand, semantic features do not seem to be important in the corpora we have used and we have corroborated again that readability features are not predictive.

Considering the cross-lingual assessment (RQ5), we see that common features have competitive results, although they do not outperform the monolingual results. This can be useful for other languages that may lack of resources as an starting point. Indeed, one of the advantages of this cross-lingual approach is that it can be easily ported to other languages. Nevertheless, these common features should be  tested in other languages to corroborate their efficacy, for example in Italian or in Portuguese.

Finally, we would like to point out the difficulty of finding corpora and datasets for this task, above all when more than two levels need to be classified. That is why we make available MultiAzterTest and all the resources at our github under GNU General Public License v3.0 (when the article is accepted). With this in mind, we would like to encourage open source work to further contribute to the field.

\section{Conclusions and future work}
\label{sec:conclusions}

In this paper, we have introduced MultiAzterTest, a multilingual analyzer of the text on multiple levels of language and Discourse-based. MultiAzterTest is an open-source tool and web service for text stylometrics and readability assessment. MultiAzterTest computes 125 indices in Basque, 141 indices in Spanish and 163 indices in English.  MultiAzterTest is mainly based on multilingual  tools, and only 24 features out of them  24 have not been obtained using universal and standard resources. The features our system analyzes texts belong to 11 levels: descriptive, syntactic complexity, word frequency, vocabulary knowledge, word information, lexical diversity, discourse connectives, referential cohesion, semantic information, semantic overlap and readability.

One of the main contributions  are the use of multilingual/universal  resources from segmentation to syntactic level using parsers based on universal dependencies, at semantic level using wordnets, at word frequency level using wordfreq, and at similarity or semantic overlap level using word embeddings. These resources are freely usable for more than 50 languages, which makes MultiAzterTest a tool that can be easily adapted.

We have tested MultiAzterTest  for readability assessment using documents written in English, Spanish and Basque, three typologically different languages.  MultiAzterTest outperforms similar tools such as Coh-Metrix, Coh-Metrix-Esp and ErreXail obtaining 90.09~\% in accuracy  when classifying into three reading levels (elementary, intermediate, and advanced) in English and  95.50~\%  in Basque and 90~\%  in Spanish when classifying into two reading levels (simple and complex) using a SMO classifier.

We also have seen that the linguistic features are easily portable across languages and a common set of features gets competitive results. The most influential feature group is the  descriptive  in the three  languages (morphology-rich as Basque and not morphology-rich as English) together with syntactic features (in second position in Basque and English and third position in Spanish).  Word information is in the third position in a morphology-rich language as Basque and sixth and fifth position in Spanish and English. Semantic information and semantic overlap features are ones that achieve the least average accuracy.

Furthermore, we have created a web application that  can be used to assess the linguistic, stylistic and readability characteristics of their reading materials in English, Spanish and Basque. And, finally, it is important to mention that all resources corpora and code\footnote{https://github.com/kepaxabier/MultiAzterTest} are publicly available as open source resources (GPL-3.0 License) in Github for reproducibility issues.


Regarding the future work, we plan to work on the adaptation of the non-exisiting resources for Basque and Spanish, e.g. the CEFR word list since the complex word identification is getting attention in the NLP community \citep{yimam-etal-2018-report}. Moreover,  we would like to keep on adding more features: more morphological features and discourse based features based on a multilingual discourse parser \citep{atutxa2019towards}.  Applying MultiAzterTest in other text classification tasks is also one of our future aims.

\section*{Acknowledgments}

We acknowledge following  projects:  DeepText (KK-2020/00088), DeepReading RTI2018-096846-B-C21 (MCIU/AEI/FEDER, UE) and BigKnowledge for Text Mining, BBVA.

%
%
\bibliographystyle{apalike} 
\bibliography{mat}

\begin{thebibliography}{68}
\providecommand{\natexlab}[1]{#1}
\providecommand{\url}[1]{\texttt{#1}}
\providecommand{\urlprefix}{}

\bibitem[{Agirrezabal et~al.(2012)Agirrezabal, Alegria, Arrieta, and
  Hulden}]{agirrezabal2012finite}
Agirrezabal, M., Alegria, I., Arrieta, B., Hulden, M.: Finite-state technology
  in a verse-making tool.
\newblock In: Proceedings of the 10th International Workshop on Finite State
  Methods and Natural Language Processing. pp. 35--39 (2012)

\bibitem[{Aguirregoitia~Martinez et~al.(2020)Aguirregoitia~Martinez,
  Bengoetxea~Kortazar, and Gonzalez-Dios}]{aguirregoitia2020clil}
Aguirregoitia~Martinez, A., Bengoetxea~Kortazar, K., Gonzalez-Dios, I.: Are
  clil texts too complicated? a computational analysis of their linguistic
  characteristics.
\newblock Journal of Immersion and Content-Based Language Education  (2020)

\bibitem[{Argamon(2019)}]{argamon2019computational}
Argamon, S.: Computational register analysis and synthesis.
\newblock Register Studies 1(1), 100--135 (2019)

\bibitem[{Atutxa et~al.(2019)Atutxa, Bengoetxea, Diaz~de Ilarraza, and
  Iruskieta}]{atutxa2019towards}
Atutxa, A., Bengoetxea, K., Diaz~de Ilarraza, A., Iruskieta, M.: Towards a
  top-down approach for an automatic discourse analysis for basque:
  Segmentation and central unit detection tool.
\newblock PloS one 14(9), e0221639 (2019)

\bibitem[{Bengoetxea and Gojenola(2010)}]{bengoetxea2010application}
Bengoetxea, K., Gojenola, K.: Application of different techniques to dependency
  parsing of basque.
\newblock In: Proceedings of the NAACL HLT 2010 First Workshop on Statistical
  Parsing of Morphologically-Rich Languages. pp. 31--39 (2010)

\bibitem[{Bengoetxea et~al.(2020)Bengoetxea, Gonzalez-Dios, and
  Aguirregoitia}]{aztertest}
Bengoetxea, K., Gonzalez-Dios, I., Aguirregoitia, A.: {AzterTest: Open Source
  Linguistic and Stylistic Analysis Tool}.
\newblock Procesamiento del Lenguaje Natural 64, 61--68 (2020)

\bibitem[{Benjamin(2012)}]{benjamin2012reconstructing}
Benjamin, R.G.: {Reconstructing readability: Recent developments and
  recommendations in the analysis of text difficulty}.
\newblock Educational Psychology Review 24(1), 63--88 (2012)

\bibitem[{Boroș et~al.(2018)Boroș, Dumitrescu, and Burtica}]{boros2018nlp}
Boroș, T., Dumitrescu, S.D., Burtica, R.: Nlp-cube: End-to-end raw text
  processing with neural networks.
\newblock In: Proceedings of the CoNLL 2018 Shared Task: Multilingual Parsing
  from Raw Text to Universal Dependencies. pp. 171--179 (2018)

\bibitem[{Brunato et~al.(2020)Brunato, Cimino, Dell’Orletta, Venturi, and
  Montemagni}]{brunato2020profiling}
Brunato, D., Cimino, A., Dell’Orletta, F., Venturi, G., Montemagni, S.:
  Profiling-ud: a tool for linguistic profiling of texts.
\newblock In: Proceedings of The 12th Language Resources and Evaluation
  Conference. pp. 7145--7151 (2020)

\bibitem[{Bryan-Legend(2006)}]{IrregularVerbsEN}
Bryan-Legend: Irregular verbs.
\newblock
  \url{https://github.com/Bryan-Legend/babel-lang/blob/master/Babel.EnglishEmitter/Resources/Irregular\%20Verbs.txt}
  (2006)

\bibitem[{Chall and Dale(1995)}]{DaleChallRevisited95}
Chall, J.S., Dale, E.: {Readability Revisited: The New Dale–Chall Readability
  Formula}.
\newblock Brookline Books, Cambridge, MA (1995)

\bibitem[{Choudhary and Arora(2020)}]{choudhary2020linguistic}
Choudhary, A., Arora, A.: Linguistic feature based learning model for fake news
  detection and classification.
\newblock Expert Systems with Applications p. 114171 (2020)

\bibitem[{Cocciu et~al.(2018)Cocciu, Brunato, Venturi, and
  Dell'Orletta}]{cocciu2018gender}
Cocciu, E., Brunato, D., Venturi, G., Dell'Orletta, F.: {Gender and Genre
  Linguistic Profiling: A Case Study on Female and Male Journalistic and Diary
  Prose}.
\newblock In: CLiC-it. pp. 131--136 (2018)

\bibitem[{Crossley et~al.(2019)Crossley, Kyle, and Dascalu}]{crossley2019tool}
Crossley, S.A., Kyle, K., Dascalu, M.: {The Tool for the Automatic Analysis of
  Cohesion 2.0: Integrating semantic similarity and text overlap}.
\newblock Behavior research methods 51(1), 14--27 (2019)

\bibitem[{da~Cunha(2015)}]{cunha2015coh}
da~Cunha, A.L.V.: Coh-Metrix-Dementia: an{\'a}lise autom{\'a}tica de
  dist{\'u}rbios de linguagem nas dem{\^e}ncias utilizando Processamento de
  L{\'i}nguas Naturais.
\newblock PhD thesis at Universidade de S{\~a}o Paulo, S{\~a}o Paulo (2015)

\bibitem[{De~Clercq and Hoste(2016)}]{de2016all}
De~Clercq, O., Hoste, V.: All mixed up? finding the optimal feature set for
  general readability prediction and its application to english and dutch.
\newblock Computational Linguistics 42(3), 457--490 (2016)

\bibitem[{Dell'Orletta et~al.(2011)Dell'Orletta, Montemagni, and
  Venturi}]{DellOrletta:2011}
Dell'Orletta, F., Montemagni, S., Venturi, G.: {READ-IT: assessing readability
  of Italian texts with a view to text simplification}.
\newblock In: Proceedings of the Second Workshop on Speech and Language
  Processing for Assistive Technologies. pp. 73--83. SLPAT '11, ACL (2011),
  \urlprefix\url{http://dl.acm.org/citation.cfm?id=2140499.2140511}

\bibitem[{Dictionaries(2020)}]{OxfordList}
Dictionaries, O.L.: Oxford 3000 and 5000: The most important words to learn in
  english.
\newblock
  \url{https://www.oxfordlearnersdictionaries.com/wordlists/oxford3000-5000}
  (2020)

\bibitem[{Eder et~al.(2016)Eder, Rybicki, and Kestemont}]{eder2016stylometry}
Eder, M., Rybicki, J., Kestemont, M.: Stylometry with r: A package for
  computational text analysis.
\newblock The R Journal 8(1) (2016)

\bibitem[{f\'acil(2020)}]{IrregularVerbsES}
f\'acil, E.: Verbos irregulares en español.
\newblock \url{https://www.esfacil.eu/es/verbos/categorias/11-irregular.html}
  (2020)

\bibitem[{Feng et~al.(2010)Feng, Jansche, Huenerfauth, and
  Elhadad}]{feng2010comparisonRedability}
Feng, L., Jansche, M., Huenerfauth, M., Elhadad, N.: {A comparison of features
  for automatic readability assessment}.
\newblock In: Proceedings of the 23rd International Conference on Computational
  Linguistics: Posters. pp. 276--284. ACL (2010)

\bibitem[{Fern{\'a}ndez~Huerta(1959)}]{fernandez1959medidas}
Fern{\'a}ndez~Huerta, J.: Medidas sencillas de lecturabilidad.
\newblock Consigna 214, 29--32 (1959)

\bibitem[{Fersini et~al.(2020)Fersini, Nozza, and
  Boifava}]{fersini2020profiling}
Fersini, E., Nozza, D., Boifava, G.: Profiling italian misogynist: An empirical
  study.
\newblock In: Proceedings of the Workshop on Resources and Techniques for User
  and Author Profiling in Abusive Language. pp. 9--13 (2020)

\bibitem[{Flesch(1979)}]{flesch1979}
Flesch, R.: {How to write plain English}.
\newblock Harper and Brothers (1979)

\bibitem[{Flesch(1948)}]{flesch1948new}
Flesch, R.: A new readability yardstick.
\newblock Journal of applied psychology 32(3), 221 (1948)

\bibitem[{Folt{\`y}nek et~al.(2019)Folt{\`y}nek, Meuschke, and
  Gipp}]{foltynek2019academic}
Folt{\`y}nek, T., Meuschke, N., Gipp, B.: Academic plagiarism detection: a
  systematic literature review.
\newblock ACM Computing Surveys (CSUR) 52(6), 1--42 (2019)

\bibitem[{Fradejas~Rueda(2020)}]{Cuentapalabras}
Fradejas~Rueda, J.M.: Cuentapalabras. Estilometr\'ia y an\'alisis de texto con
  R para fil\'ologos.
\newblock http://www.aic.uva.es/cuentapalabras/ (2020)

\bibitem[{Fran{\c{c}}ois and Fairon(2012)}]{franccois2012ai}
Fran{\c{c}}ois, T., Fairon, C.: {An AI readability formula for French as a
  foreign language}.
\newblock In: Proceedings of the 2012 Joint Conference on Empirical Methods in
  Natural Language Processing and Computational Natural Language Learning. pp.
  466--477. ACL (2012)

\bibitem[{Gonzalez-Agirre et~al.(2012)Gonzalez-Agirre, Laparra, and
  Rigau}]{gonzalez2012multilingual}
Gonzalez-Agirre, A., Laparra, E., Rigau, G.: Multilingual central repository
  version 3.0.
\newblock In: LREC. pp. 2525--2529 (2012)

\bibitem[{Gonzalez-Dios et~al.(2014)Gonzalez-Dios, Aranzabe, D{\'\i}az~de
  Ilarraza, and Salaberri}]{errexail}
Gonzalez-Dios, I., Aranzabe, M.J., D{\'\i}az~de Ilarraza, A., Salaberri, H.:
  Simple or complex? assessing the readability of basque texts.
\newblock In: Proceedings of {COLING} 2014, the 25th International Conference
  on Computational Linguistics: Technical Papers. pp. 334--344. DCU and ACL,
  Dublin, Ireland (Aug 2014),
  \urlprefix\url{https://www.aclweb.org/anthology/C14-1033}

\bibitem[{Gonzalez-Dios et~al.(2020)Gonzalez-Dios, Bengoetxea, and
  Aguirregoitia}]{laguntest}
Gonzalez-Dios, I., Bengoetxea, K., Aguirregoitia, A.: {LagunTest: A NLP Based
  Application to Enhance Reading Comprehension}.
\newblock In: 1st Workshop on Tools and Resources to Empower People with
  REAding DIfficulties (READI2020). pp. 63--69 (2020)

\bibitem[{Graesser et~al.(2011)Graesser, McNamara, and
  Kulikowich}]{graesser2011cohMetrix}
Graesser, A.C., McNamara, D.S., Kulikowich, J.M.: {Coh-Metrix Providing
  Multilevel Analyses of Text Characteristics}.
\newblock Educational Researcher 40(5), 223--234 (2011)

\bibitem[{Graesser et~al.(2004)Graesser, McNamara, Louwerse, and
  Cai}]{graesser2004coh}
Graesser, A.C., McNamara, D.S., Louwerse, M.M., Cai, Z.: Coh-metrix: Analysis
  of text on cohesion and language.
\newblock Behavior research methods, instruments, \& computers 36(2), 193--202
  (2004)

\bibitem[{Gunning(1968)}]{gunning1968technique}
Gunning, R.: {The technique of clear writing}.
\newblock McGraw-Hill New York (1968)

\bibitem[{Hall et~al.(2009)Hall, Frank, Holmes, Pfahringer, Reutemann, and
  Witten}]{hall2009weka}
Hall, M., Frank, E., Holmes, G., Pfahringer, B., Reutemann, P., Witten, I.H.:
  {The WEKA data mining software: an update}.
\newblock ACM SIGKDD Explorations Newsletter 11(1), 10--18 (2009)

\bibitem[{Hancke et~al.(2012)Hancke, Vajjala, and
  Meurers}]{hanckeReadabilityGerman12}
Hancke, J., Vajjala, S., Meurers, D.: {Readability Classification for German
  using lexical, syntactic, and morphological features}.
\newblock In: COLING 2012: Technical Papers. p. 1063–1080 (2012)

\bibitem[{Hou and Huang(2020)}]{hou2020robust}
Hou, R., Huang, C.R.: Robust stylometric analysis and author attribution based
  on tones and rimes.
\newblock Natural Language Engineering 26(1), 49--71 (2020)

\bibitem[{ISO(2020)}]{stopwords}
ISO, S.: Stopwords iso: The most comprehensive collection of stopwords for
  multiple languages.
\newblock \url{https://github.com/stopwords-iso/stopwords-iso} (2020)

\bibitem[{Kyle et~al.(2018)Kyle, Crossley, and Berger}]{kyle2018tool}
Kyle, K., Crossley, S., Berger, C.: {The tool for the automatic analysis of
  lexical sophistication (TAALES): version 2.0}.
\newblock Behavior research methods 50(3), 1030--1046 (2018)

\bibitem[{Madrazo and Pera(2019)}]{madrazo19}
Madrazo, I., Pera, M.S.: Multiattentive recurrent neural network architecture
  for multilingual readability assessment.
\newblock Transactions of the Association for Computational Linguistics 7,
  421--436 (2019)

\bibitem[{Madrazo~Azpiazu and Pera(2020)}]{madrazo2020cross}
Madrazo~Azpiazu, I., Pera, M.S.: Is cross-lingual readability assessment
  possible?
\newblock Journal of the Association for Information Science and Technology
  71(6), 644--656 (2020)

\bibitem[{Martin-Borregon(2014)}]{esSilaba}
Martin-Borregon, D.: Sibilizador.
\newblock \url{https://github.com/mabodo/sibilizador} (2014)

\bibitem[{Mc~Laughlin(1969)}]{mc1969smog}
Mc~Laughlin, G.H.: {SMOG grading-a new readability formula}.
\newblock Journal of reading 12(8), 639--646 (1969)

\bibitem[{McCarthy(2005)}]{mccarthy2005assessment}
McCarthy, P.M.: An assessment of the range and usefulness of lexical diversity
  measures and the potential of the measure of textual, lexical diversity
  (MTLD).
\newblock PhD thesis at The University of Memphis, Memphis (2005)

\bibitem[{McLaughlin(1969)}]{mclaughlin1969clearing}
McLaughlin, G.H.: Clearing the smog.
\newblock J Reading  (1969)

\bibitem[{Melka and M{\'\i}steck{\`y}(2019)}]{melka2019stylometric}
Melka, T.S., M{\'\i}steck{\`y}, M.: {On Stylometric Features of H. Beam
  Piper’s Omnilingual}.
\newblock Journal of Quantitative Linguistics pp. 1--40 (2019)

\bibitem[{Mikolov et~al.(2018)Mikolov, Grave, Bojanowski, Puhrsch, and
  Joulin}]{mikolov2018advances}
Mikolov, T., Grave, E., Bojanowski, P., Puhrsch, C., Joulin, A.: Advances in
  pre-training distributed word representations.
\newblock In: Proceedings of the International Conference on Language Resources
  and Evaluation (LREC 2018). pp. 52--55 (2018)

\bibitem[{Miller(1995)}]{miller1995wordnet}
Miller, G.A.: Wordnet: a lexical database for english.
\newblock Communications of the ACM 38(11), 39--41 (1995)

\bibitem[{Miro{\'n}czuk and Protasiewicz(2018)}]{mironczuk2018recent}
Miro{\'n}czuk, M.M., Protasiewicz, J.: A recent overview of the
  state-of-the-art elements of text classification.
\newblock Expert Systems with Applications 106, 36--54 (2018)

\bibitem[{Ortiz-Zambranoa and Montejo-R{\'a}ezb(2020)}]{ortiz2020overview}
Ortiz-Zambranoa, J.A., Montejo-R{\'a}ezb, A.: Overview of alexs 2020: First
  workshop on lexical analysis at sepln.
\newblock In: Proceedings of the Iberian Languages Evaluation Forum (IberLEF
  2020) (2020)

\bibitem[{Padr{\'o} et~al.(2010)Padr{\'o}, Collado, Reese, Lloberes, and
  Castell{\'o}n}]{padro2010freeling}
Padr{\'o}, L., Collado, M., Reese, S., Lloberes, M., Castell{\'o}n, I.:
  Freeling 2.1: Five years of open-source language processing tools.
\newblock In: 7th International Conference on Language Resources and
  Evaluation. pp. 931--936 (2010)

\bibitem[{Paetzold and Specia(2016)}]{paetzold2016semeval}
Paetzold, G., Specia, L.: {Semeval 2016 task 11: Complex word identification}.
\newblock In: Proceedings of the 10th International Workshop on Semantic
  Evaluation (SemEval-2016). pp. 560--569 (2016)

\bibitem[{Petersen and Ostendorf(2009)}]{petersen2009machine}
Petersen, S.E., Ostendorf, M.: {A machine learning approach to reading level
  assessment}.
\newblock Computer Speech \& Language 23(1), 89--106 (2009)

\bibitem[{Platt(1998)}]{platt1998sequential}
Platt, J.: Sequential minimal optimization: A fast algorithm for training
  support vector machines.
\newblock Tech. Rep. MSR-TR-98-14 (1998)

\bibitem[{Pociello et~al.(2011)Pociello, Agirre, and
  Aldezabal}]{pociello2011methodology}
Pociello, E., Agirre, E., Aldezabal, I.: Methodology and construction of the
  basque wordnet.
\newblock Language resources and evaluation 45(2), 121--142 (2011)

\bibitem[{Qi et~al.(2019)Qi, Dozat, Zhang, and Manning}]{qi2019universal}
Qi, P., Dozat, T., Zhang, Y., Manning, C.D.: Universal dependency parsing from
  scratch.
\newblock arXiv preprint arXiv:1901.10457  (2019)

\bibitem[{Quispesaravia et~al.(2016)Quispesaravia, Perez, Cabezudo, and
  Alva-Manchego}]{cohmetrixES}
Quispesaravia, A., Perez, W., Cabezudo, M.S., Alva-Manchego, F.:
  {Coh-Metrix-Esp: A complexity analysis tool for documents written in
  Spanish}.
\newblock In: Proceedings of the Tenth International Conference on Language
  Resources and Evaluation (LREC'16). pp. 4694--4698 (2016)

\bibitem[{Scarton and Alu{\i}sio(2010)}]{scarton2010coh}
Scarton, C., Alu{\i}sio, S.M.: Coh-metrix-port: a readability assessment tool
  for texts in brazilian portuguese.
\newblock In: Proceedings of the 9th International Conference on Computational
  Processing of the Portuguese Language, Extended Activities Proceedings,
  PROPOR. vol.~10, pp. 1--2 (2010)

\bibitem[{Scarton and Alu{\'\i}sio(2010)}]{scarton2010analise}
Scarton, C.E., Alu{\'\i}sio, S.M.: An{\'a}lise da inteligibilidade de textos
  via ferramentas de processamento de l{\'\i}ngua natural: adaptando as
  m{\'e}tricas do coh-metrix para o portugu{\^e}s.
\newblock Linguam{\'a}tica 2(1), 45--61 (2010)

\bibitem[{Schicchi et~al.(2020)Schicchi, Pilato, and Bosco}]{schicchi2020deep}
Schicchi, D., Pilato, G., Bosco, G.L.: Deep neural attention-based model for
  the evaluation of italian sentences complexity.
\newblock In: 2020 IEEE 14th International Conference on Semantic Computing
  (ICSC). pp. 253--256. IEEE (2020)

\bibitem[{Si and Callan(2001)}]{si2001statistical}
Si, L., Callan, J.: {A statistical model for scientific readability}.
\newblock In: Proceedings of the tenth international conference on Information
  and knowledge management. pp. 574--576. ACM (2001)

\bibitem[{Speer et~al.(2018)Speer, Chin, Lin, Jewett, and Nathan}]{robyn-speer}
Speer, R., Chin, J., Lin, A., Jewett, S., Nathan, L.: Luminosoinsight/wordfreq:
  v2.2 (Oct 2018), \urlprefix\url{https://doi.org/10.5281/zenodo.1443582}

\bibitem[{Steven and Edward(2006)}]{steven2006nltk}
Steven, B., Edward, L.: Nltk: the natural language toolkit.
\newblock In: Proceedings of the COLING/ACL on Interactive presentation
  sessions. pp. 69--72 (2006)

\bibitem[{Vajjala and Lu{\v{c}}i{\'c}(2018)}]{vajjala2018onestopenglish}
Vajjala, S., Lu{\v{c}}i{\'c}, I.: Onestopenglish corpus: A new corpus for
  automatic readability assessment and text simplification.
\newblock In: Proceedings of the thirteenth workshop on innovative use of NLP
  for building educational applications. pp. 297--304 (2018)

\bibitem[{Weide(2005)}]{weide2005carnegie}
Weide, R.: The carnegie mellon pronouncing dictionary [cmudict. 0.6] (2005)

\bibitem[{Willits et~al.(2018)Willits, Rubin, Jones, Minor, and
  Lysaker}]{willits2018evidence}
Willits, J.A., Rubin, T., Jones, M.N., Minor, K.S., Lysaker, P.H.: Evidence of
  disturbances of deep levels of semantic cohesion within personal narratives
  in schizophrenia.
\newblock Schizophrenia research 197, 365--369 (2018)

\bibitem[{Yimam et~al.(2018)Yimam, Biemann, Malmasi, Paetzold, Specia,
  {\v{S}}tajner, Tack, and Zampieri}]{yimam-etal-2018-report}
Yimam, S.M., Biemann, C., Malmasi, S., Paetzold, G., Specia, L., {\v{S}}tajner,
  S., Tack, A., Zampieri, M.: A report on the complex word identification
  shared task 2018.
\newblock In: Proceedings of the Thirteenth Workshop on Innovative Use of {NLP}
  for Building Educational Applications. pp. 66--78. Association for
  Computational Linguistics, New Orleans, Louisiana (Jun 2018),
  \urlprefix\url{https://www.aclweb.org/anthology/W18-0507}

\bibitem[{Zeman and Haji{\v{c}}(2018)}]{K18-2:2018}
Zeman, D., Haji{\v{c}}, J. (eds.): Proceedings of the {CoNLL} 2018 Shared Task:
  Multilingual Parsing from Raw Text to Universal Dependencies.
\newblock ACL, Brussels, Belgium (October 2018),
  \urlprefix\url{http://www.aclweb.org/anthology/K18-2}

\end{thebibliography}
\end{document}